\documentclass[10pt, a4paper]{article}

\usepackage[english]{babel}
\usepackage{lipsum} 
\usepackage[letterpaper,top=2cm,bottom=2cm,left=3cm,right=3cm,marginparwidth=1.75cm]{geometry}
\usepackage[final]{lrec-coling2024} 
\renewcommand{\thefootnote}{\arabic {각주}}
\usepackage{multirow}
\usepackage{tabularx}
\usepackage{adjustbox}
\usepackage{kotex} 
\usepackage{amsmath}
\usepackage{amssymb}
\usepackage{graphicx} 
\usepackage{array}    
\usepackage{verbatim}
\usepackage{tikz}
\usepackage{graphicx}
\usepackage{caption}
\usepackage{booktabs}

\title{A New Korean Text Classification Benchmark for \\ Recognizing the Political Intents in Online Newspapers}

\name{Beomjune Kim$^\ast$ \; Eunsun Lee$^\ast$ \; Dongbin Na$^\dagger$\thanks{$^\dagger$Correspondence to dongbinna@postech.ac.kr}} 
\address{SEOULTECH \; Kyung Hee University (KHU) \; POSTECH}

\abstract{
Many users reading online articles in various magazines may suffer considerable difficulty in distinguishing the implicit intents in texts.
In this work, we focus on automatically recognizing the political intents of a given online newspaper by understanding the context of the text. To solve this task, we present a novel Korean text classification dataset that contains various articles. We also provide deep-learning-based text classification baseline models trained on the proposed dataset.
Our dataset contains 12,000 news articles that may contain political intentions, from the \textit{politics} section of six of the most representative newspaper organizations in South Korea.
All the text samples are labeled simultaneously in two aspects (1) the level of political orientation and (2) the level of pro-government.
To the best of our knowledge, our paper is the most large-scale Korean news dataset that contains long text and addresses multi-task classification problems. We also train recent state-of-the-art (SOTA) language models that are based on transformer architectures and demonstrate that the trained models show decent text classification performance.
All the codes, datasets, and trained models are available at \textbf{\textcolor{pink}{\url{https://github.com/Kdavid2355/KoPolitic-Benchmark-Dataset}}}. \\ \newline }

\begin{document}

\maketitleabstract

\def\thefootnote{*}\footnotetext{These authors contributed equally to this work.}

\section{Introduction}
As online social services develop, the appearance of online texts has become more diverse and complex. Therefore, the ability to distinguish informative content for users has been unprecedentedly needed. However, users are facing difficulties in making choices in the face of vast amounts of information.
We encounter selective information from recommendation algorithms, which might reduce the diversity of information and sometimes make users biased~\cite{bias1, bias2, filterbubble_book1,filterbubble_book2}. Although articles tend to contain relatively unbiased information compared to other media, articles can also contain biased information depending on the diverse reasons~\cite{media_bias_2010, media_bais_2019}. In this work, we study recognizing the political intent and government criticism of political articles using deep-learning models.

Previous work has utilized various machine learning algorithms to solve political-related classification tasks such as political orientation classification~\cite{political_orientation, twitter_cls1, twitter_cls3} and election prediction~\cite{predict-election, us_brazil_elction_prediction}.
We note that most existing work has predominantly focused on single classification tasks using short texts, such as those from Twitter, to classify political orientations.
However, in the real world, text data frequently contains a long sequence of words, including various intentions and semantic information.
For instance, a single article might not only convey a political intent but also embed implicit pro-government sentiments.
Therefore, we propose a new dataset that contains real-world online newspapers with \textit{each article's (1) political orientation and (2) the level of pro-government.}
This dataset classifies the political orientation (liberal or conservative tendencies) of an article using a 5-point scale and evaluates the level of pro-government using a 6-point scale, including an additional \textit{None} category (See Table~\ref{fig:dataset_configuration}).

We note that our proposed dataset provides a lot of text data that contains diverse political information.
To validate the usefulness of our proposed dataset, we leverage state-of-the-art transformer architectures, specifically KoBERT~\cite{kobert}, KoBigBird~\cite{kobigbird}, and KoELECTRA~\cite{koelectra}.
We fine-tune the pre-trained models on our proposed dataset to solve the two different classification problems.
As a result, we demonstrate that transformer-based architectures utilizing attention mechanisms can be useful in recognizing the whole semantic information of the long text.

\begin{figure*}[htp]
    \centering   
    \centerline{\includegraphics[width=1.10\textwidth]{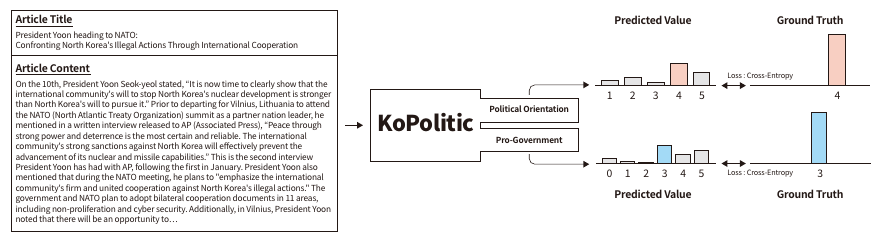}}
    \caption{The conceptual illustration of our proposed model with the input and output examples. We note that our proposed architecture simultaneously predicts two types of outputs, (1) political orientation and (2) the level of attitude towards government by utilizing multi-task learning loss.}
    \label{fig:model_configuration}
\end{figure*}

In particular, we propose our multi-task model, \textbf{KoPolitic}, which has been trained on our dataset by leveraging the fine-tuning based on the recent KoBigBird~\cite{kobigbird} architecture.
Our proposed text classification model is capable of simultaneously classifying political intent and the level of pro-government.
A detailed overview of our dataset and methodology is depicted in Figure~\ref{fig:model_configuration}.
Our approach to handling two distinct tasks concurrently provides an effective solution that facilitates a more granular and accurate analysis of articles from diverse perspectives.

Our main contributions are as follows:

\begin{itemize}
\item We introduce a novel multi-task Korean text classification dataset designed for recognizing long text data that might contain political intents. This dataset contains extensive annotations including political inclinations and the government's affinity level.
\item Our proposed model, \textbf{KoPolitic}, effectively solves multiple tasks simultaneously, reporting improved classification performance by leveraging the advantage of multi-task learning in each classification task compared to the other baseline benchmarks.
\item We publicly provide all resources, including the dataset and source code, for academic research and real-world applications.

\end{itemize}

\section{Background}

The Natural Language Processing (NLP) research fields can be categorized into two types of approaches, conventional statistical-based methods and neural network-based methods. The traditional statistical-based models generally assign probabilities to sequences directly. In particular, N-gram models predict the likelihood of a given sequence utilizing the relationship between neighboring words and a word. However, recently, language models have mainly adopted neural network architectures due to the sub-par classification performance and sparsity problems of traditional methods. Recent deep learning-based language models have been broadly adopted in various NLP applications. Since the concept of the attention mechanism has emerged~\cite{attention}, NLP models perform better in addressing longer or more complex sentences. Especially, recent self-attention-based transformers process all input tokens simultaneously in the forward process and allow hardware accelerators to efficiently process large datasets. This architectural innovation has led to the growth of large-scale language models such as BERT~\cite{bert} and GPT~\cite{attention, transformer, bert, gpt}.

The recent BERT-based models can understand the large amounts of text data and improve the classification performance by leveraging a pre-training process and the subsequent fine-tuning~\cite{bert}. Moreover, many derivatives of BERT have been proposed such as BigBird~\cite{bigbird} and ELECTRA~\cite{electra}. The naive BERT-based architectures have a limitation in handling long sequences because the amount of computation increases proportionally to the square of the sequence length. To remedy this computational burden of BERT, some studies have proposed breakthrough methods. BigBird~\cite{bigbird} leverages a sparse attention mechanism instead of full attention in the self-attention layer. The sparse attention mechanism is described as having the properties of random attention and window attention, and furthermore, BigBird improves the generalization performance by leveraging the global attention that actively utilizes global tokens. On the other hand, ELECTRA adopts both discriminator and generator architectures, which can result in improved classification performance~\cite{electra} compared to the conventional BERT architecture~\cite{bert}.

\section{Related Works}

\subsection{Political Text Recognition}

Previous studies on political text recognition have utilized various types of data such as speech texts~\cite{speech2, political_speech_party, speech1-2019, speech3-2013}, social media data~\cite{predict-election, china, it's_not_easy, twitter_cls1, twitter_cls3, twitter_cls2, Beyond-Binary-Labels-2017} and article texts~\cite{2016-political-Sentiment, political-article}. 
In particular, various work has studied political corpora to predict the political orientation of users~\cite{it's_not_easy, political_orientation, twitter_cls1, twitter_cls3}.

Among them, unlike previous work using only binary labels (true or false), a foundation study has initially presented a new dataset, in which text data belongs to one category among seven classes by adopting a multi-class approach for recognizing the political orientation. The work has focused on predicting political orientation given \textit{short} speech texts collected from Twitter and analyzed various aspects including the frequency of words. The work has shown that predicting subtle political ideology in the real world could be more difficult than previous problems only addressing binary labels~\cite{Beyond-Binary-Labels-2017}. 
Similarly, we adopt the multi-class annotations, however, our work also provides a level of pro-government additionally per each text and also contains long texts.
Moreover, other studies have used political corpora to predict the future of diplomatic relations~\cite{china} or to analyze the sentiment of political posts on social media~\cite{2016-political-Sentiment, election-2011} and predict elections~\cite{predict-election, us_brazil_elction_prediction, election-2011}.

\subsection{News Article Classification}

Although articles fundamentally are expected to convey objective facts, journalists sometimes reveal their own intentions~\cite{media_bias_2010, media_bais_2019, news_bais_2000} for a specific purpose. Sentiment analysis is a useful tool to predict whether a text contains positive, negative, or neutral connotations. In particular, sentiment analysis for opinions of online users on various political issues can help understand their political activities. Previous studies have attempted sentiment analysis~\cite{sentiment-2013, article-sentiment, article-sentiment2, news-Sentiment-2020} of news articles. Especially, a previous study utilizing political articles also has conducted sentiment analysis, addressing the way to analyze sentiment even from sparse sources~\cite{2016-political-Sentiment}. Another work has utilized not only the content of the article but also the title and link structure to predict political orientation~\cite{multi-label-political-cls}. The previous work that analyzes political orientation or sentiments of political articles generally utilizes only 3 labels~\cite{2016-political-Sentiment, multi-label-political-cls}. However, compared to the previous studies, we adopt 5 labels to detect more subtle political ideology and, furthermore, analyze political orientation and sentiments comprehensively.

\section{Proposed Dataset}

Our dataset consists of 12,000 news articles from the \textit{politics} section that have been collected from various online Korean magazines, of which 5,000 have been directly collected and labeled by human labelers. The remaining 7,000 articles are unlabeled and can be used for unsupervised learning. The labeled dataset includes politically biased articles, those leaning towards or against the government. We note that our text samples mainly consist of more than 500-word tokens. Therefore, our proposed dataset requires comprehensively leveraging the feature representations across the long sentences~\cite{bigbird}. Article texts of our dataset are broadly categorized based on their level of political orientation, which are labeled on a scale from 1 to 5. Additionally, these articles are also concurrently labeled based on their criticism or advocacy towards the government and also annotated on a scale from 0 to 5. Unlike the criteria for political orientation, the pro-government criterion considers text instances where the government, the subject of the advocacy or criticism, is not mentioned. Due to this reason, we additionally have adopted a label of 0, which denotes \textit{None} on the second criterion.

\subsection{Data Collection Pipeline}

Social media companies that publish various news articles are sometimes associated with a specific political party. To reduce the label imbalance, we have curated our dataset by consistently crawling 2,000 articles each from a total of 6 different magazines. Specifically, we have adopted two of the most representative conservative-leaning newspapers, two liberal-leaning newspapers, and two moderate-leaning newspapers in South Korea.

\begin{table*}
\centering
\begin{tabular}{c|c|c|c|c|c|c} 
\hline
\textbf{Class} & 0 & 1   & 2  & 3  & 4  & 5  \\ 
\hline
\begin{tabular}[c]{@{}c@{}}\textbf{Political} \\ \textbf{Orientation}\end{tabular}         
& - & Liberal   & \begin{tabular}[c]{@{}c@{}}Moderate\\Liberal\end{tabular}   & Moderate & \begin{tabular}[c]{@{}c@{}}Moderate\\Conservative\end{tabular} & Conservative \\ 
\hline
\begin{tabular}[c]{@{}c@{}}\textbf{The Level of} \\ \textbf{Pro-government}\end{tabular} & None & Criticism & \begin{tabular}[c]{@{}c@{}}Moderate\\Criticism\end{tabular} & Moderate & \begin{tabular}[c]{@{}c@{}}Moderate\\Advocacy\end{tabular}     & Advocacy   \\
\hline
\end{tabular}
\caption{Classification criteria of our proposed dataset. Our dataset provides two labels per text that can be used for multi-task learning. The categories (classes) are ordinally composed in the two tasks (1) recognizing political orientation and (2) classifying the level of pro-goverment of the articles.}
\label{fig:dataset_configuration}
\end{table*}

\subsection{Data Labeling Process}

A team that is composed of five annotators has conducted the data labeling process.
During the annotation phase, when encountering ambiguous or unclear articles, we have labeled that particular article after a discussion among the five annotators.
In cases of disagreement, a pre-appointed moderator has decided on the final annotation.
After completing the labeling phase, we have conducted a cross-check among the annotators to ensure the originality of the data distribution and to minimize label imbalance.

\begin{figure}[htp]
    \centering
    \includegraphics[width=\columnwidth]{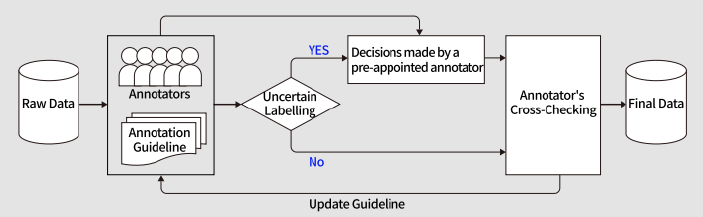}
    \caption{Overview of the data labeling process.}
    \label{fig:datset_configuration2}
\end{figure}

\noindent The criteria and considerations for the labeling process are as follows:

\begin{itemize}

\item \textbf{Classifying Political Intent}

The labelers have labeled the articles based on the journalist's intent.
The articles receive a class of '1' for liberal intent, '2' for moderate liberal intent, '3' for neutral presentation or straightforward information relay, '4' for moderate conservative intent, and '5' for conservative intent.
When the journalist's intention of an article is unclear, we label the article text considering the traditional values pursued by the political party.

\item \textbf{Determining the Level of Pro-Government}

If an article shows a particularly strong critical tone or condemnation, the labelers have given the text a class of 1. Articles that have a critical tone towards the government have received a class of 2. Articles detailing the government's role, events, or ceremonies are given a class of 3. When an article is complimentary or favorable towards the government, the article sample has received a class of 4. If an article shows strong admiration or praise for the government, the labelers have given the article a class of 5. When the government is not the subject of discussion at all, the article has been labeled with a class of 0.

\end{itemize}

\subsection{Analysis of Dataset}

After collecting and annotating the article samples, we have found that labeled 5,000 articles show an imbalanced data distribution across various distinct classes. We have avoided any additional selection processes to maintain the true distribution in a real-world setting for constructing the training dataset, which tends to induce class imbalance. Figure~\ref{fig:datset_configuration2} shows that the distribution of the training dataset dominantly contains \textit{moderate} text samples with a class of 3 for both the level of political orientation and pro-government criteria.

\begin{figure}[htp]
    \centering
    \includegraphics[width=\columnwidth]{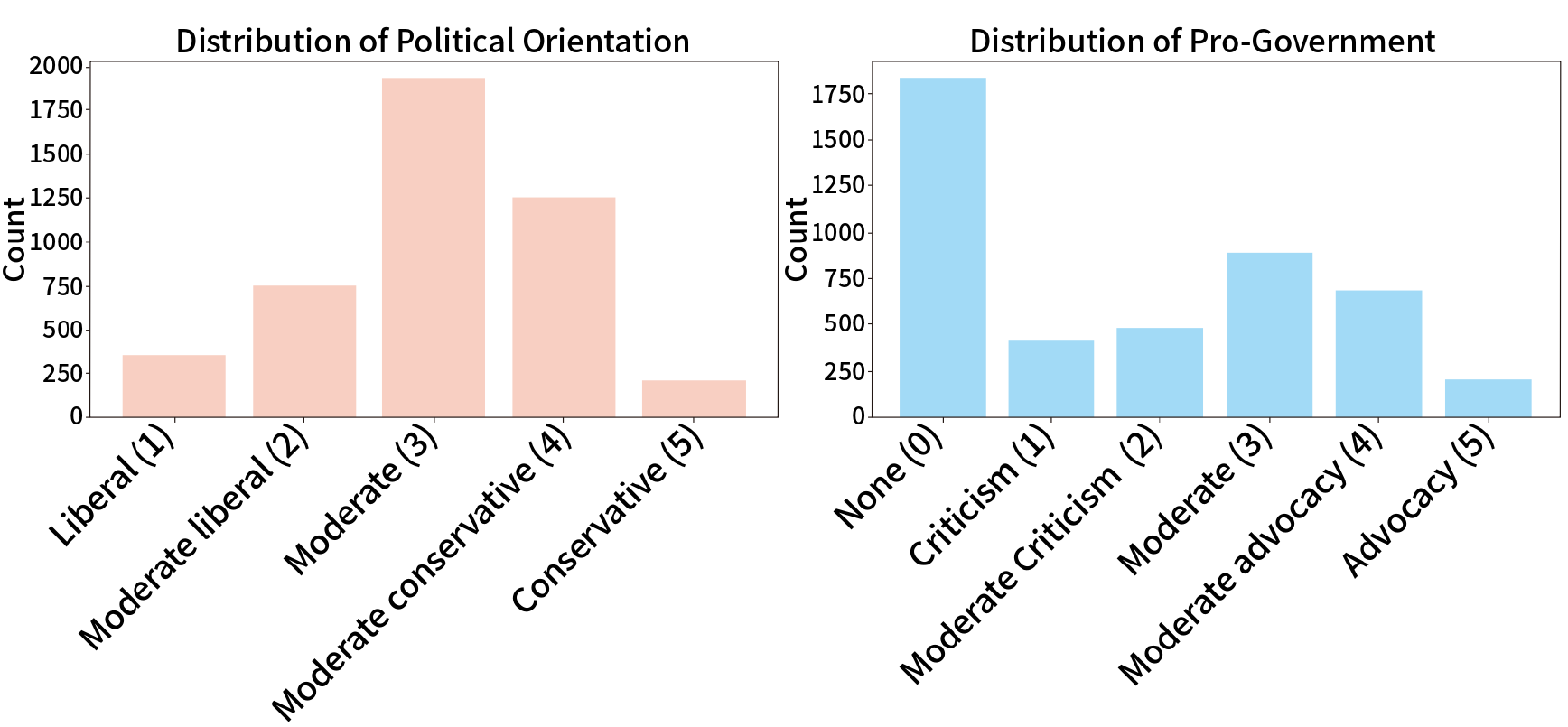}
    \caption{Class distribution of the training dataset. This figure represents the number of text samples across categories: the class distribution of (1) degrees of political orientation, and (2) the level of pro-government in our training dataset.}
    \label{fig:datset_configuration2}
\end{figure}

\begin{figure}[htp]
    \centering
    \includegraphics[width=\columnwidth]{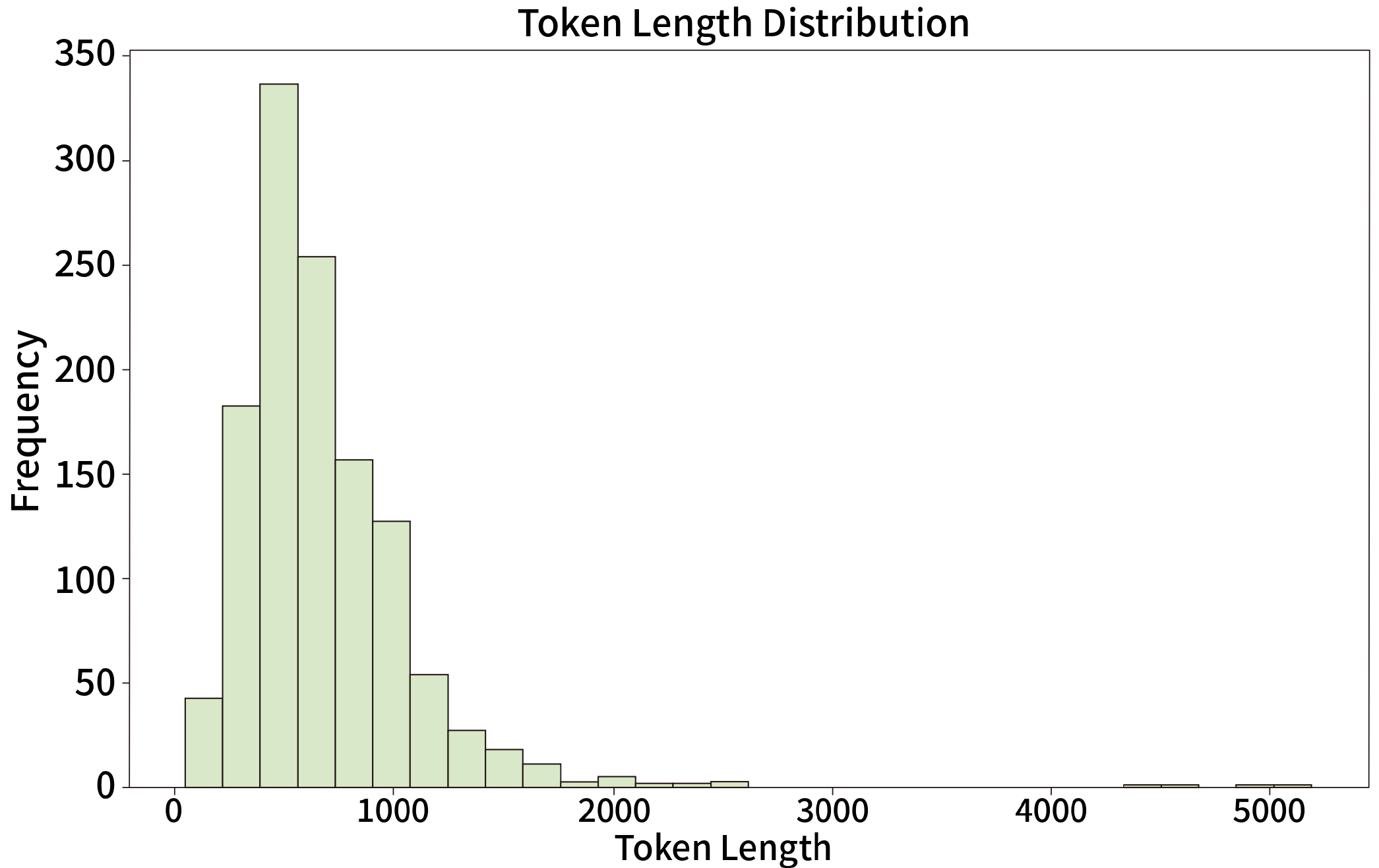}
    \caption{Distribution of the number of tokens in the text samples in the training dataset. Most text samples usually contain about 500 tokens as shown in the distribution.}
    \label{fig:datset_token}
\end{figure}
\vspace{-1pt}
\section{Proposed Architecture}

We present a multi-task model, \textbf{KoPolitic} that is greatly effective for comprehensively recognizing the political intents of article texts. Our model uses texts as input data $x$. Given our proposed dataset $\mathcal{D} \in\{\mathcal{D}_{train}, \mathcal{D}_{test}\}$, where $\mathcal{D}_{train} = \{x_i, y_i\}_{i=1}^N$ and $N$ denotes the total number of training data samples, our model aims to predict the political intents of a text sample $x$. $x_i$ represents the $i$-th input text data, and $y_i$ is the corresponding target (class) to $x_i$. Thus, our problem can be defined by the following maximum log-likelihood estimation formulation:

\begin{equation}
    \theta^{*} = \arg \max_{\theta} \log p(y_i | x_i; \theta)
\end{equation}

\noindent where $\theta$ represents the set of model weight parameters and the $\theta^*$ denotes the ideal optimal parameters of the probabilistic model. In detail, the $y_i$ consists of two annotations $\{ y^{1}_i, y^{2}_i \}$ in our proposed dataset. The $y^{1}_i$ denotes the political orientation label across 5 categories and $y^{2}_i$ represents the pro-government degree across 6 categories for the $i$-th text data $x_i$. The classification model $\theta$ solves the two individual tasks simultaneously given the input data $x_i$ in a multi-task manner. Figure~\ref{fig:model_configuration} provides a conceptual illustration of the model. To train weights of the classifier $\theta$, our objective is to minimize the cross-entropy loss. Formally, our classification heads $H_1(\cdot)$ and $H_2(\cdot)$ solve two tasks respectively. The feature representation $v=F(x)$ is extracted using the foundation language model network $F$. Therefore, our final loss function is formulated as follows:

\begin{equation}
    \mathbb{E}_{(x, y^{1}, y^{2}) \in \mathcal{D}_{train}}[l(H_1(v), y^1) + \lambda \cdot l(H_2(v), y^2)]
\end{equation}

\begin{table*}
\centering
\renewcommand\arraystretch{1.0}
\begin{adjustbox}{width=\textwidth,center}
\begin{tabular}{c|c|c|c|c|c|c|c|c}
\hline
\multicolumn{3}{c|}{Training Models} & \multicolumn{6}{c}{Test Dataset Results}    \\ 
\hline
\multicolumn{2}{c|}{\multirow{2}{*}{Architecture}} & \multirow{2}{*}{Seq Len} & \multicolumn{3}{c|}{\textbf{Political Orientation}} & \multicolumn{3}{c}{\textbf{The Level of Pro-government}} \\
\cline{4-9}
\multicolumn{2}{c|}{} & & \textbf{Top-2 Accuracy} & \textbf{Accuracy} & \textbf{F1-Score} & \textbf{Top-2 Accuracy} & \textbf{Accuracy} & \textbf{F1-Score} \\ 
\hline
\multirow{3}{*}{KoBigBird} & \multirow{3}{*}{Single} & 1024 & 0.7720 & 0.4900 & 0.4612 & 0.8040 & 0.6220 & 0.5500 \\
 &  & 2048 & 0.7240 & 0.4720 & 0.4417 & 0.7320 & 0.5780 & 0.4852 \\
 &  & 3072 & 0.7020 & 0.4560 & 0.4263 & 0.7820 & 0.6040 & 0.5229 \\
\hline
\multirow{3}{*}{KoPolitic (\textbf{Ours})} & \multirow{3}{*}{Multi} & 1024 & 0.7080 & 0.4380 & 0.3828 & 0.7620 & 0.5980 & 0.5179 \\
 &  & 2048 & 0.7260 & 0.4460 & 0.3938 & 0.7540 & 0.5800 & 0.4917 \\
 &  & 3072 & 0.7560 & 0.4820 & 0.4734 & 0.7840 & 0.6120 & 0.5455 \\
\hline
\multirow{2}{*}{KoBERT} & Single & 512 & 0.6360 & 0.3840 & 0.3617 & 0.7020 & 0.5360 & 0.4678 \\
 & Multi & 512 & 0.5620 & 0.3480 & 0.2905 & 0.7080 & 0.4940 & 0.3451 \\
\hline
\multirow{2}{*}{KoElectra} & Single & 512 & 0.7620 & 0.5120 & 0.4949 & 0.7420 & 0.5700 & 0.4454 \\
 & Multi & 512 & 0.7340 & 0.4660 & 0.4487 & 0.7640 & 0.6020 & 0.5356 \\
\hline
\end{tabular}

\end{adjustbox}

\caption{Classification performance comparison results for political orientation and level of government tasks, using metrics such as Top-2 accuracy, accuracy, and F1-macro score. The higher is better.}

\label{tab:result_table}
\end{table*}

\begin{table}[h]
\centering
\large
\renewcommand{\arraystretch}{1.3} 
\begin{adjustbox}{width=0.45\textwidth}
\begin{tabular}{l|cc|cc}
\hline
& \multicolumn{2}{c|}{Political Orientation} & \multicolumn{2}{c}{Pro-Government} \\
\cline{2-5}
\multirow{-2}{*}{Multi-Model} & MAE & Hamming Loss & MAE & Hamming Loss \\
\hline
KoBigBird (1024) & 0.7320 & 0.5680 & 0.7860 & 0.4300 \\
\hline
\textbf{KoPolitic (3072)}      & 0.6952 & 0.5159 & 0.7410 & 0.3864 \\
\hline
KoBERT          &1.1060 & 0.6520 & 1.1020 & 0.5060 \\
KoELECTRA       &0.7260 &0.5340 &0.7840 & 0.3980 \\
\hline
\end{tabular}
\end{adjustbox}
\caption{Performance comparison results using MAE (Mean Absolute Error) and Hamming loss. The lower is better.}
\label{loss}
\end{table}
\vspace{-8pt}

\noindent For the foundation models, we utilize three popular BERT-based models: KoBERT~\cite{kobert}, KoELECTRA~\cite{koelectra}, and KoBigBird~\cite{kobigbird}. For comparing various model architectures, we have also trained the single-task models that solely classify just one task among the political orientation classification or the level of pro-government classification. We note that the required computational memory and inference time increases in proportion to the number of tasks when using the individual single-task models. In contrast, our multi-head model \textbf{KoPolitic} that solves the two different tasks simultaneously for a single text sample requires lower memory resources and inference time, by approximately 2 times, compared to the individual single-task models that only address an individual task.
These improvements in complexity are attributed to the desirable property of the multi-task architectures.
The multi-task model can extract feature vectors by forwarding the input texts into the feature extractor network $F(\cdot)$ only once for solving multiple different tasks~\cite{kim2023problem,choi2023large}.
For training the models, we have utilized either Adam~\cite{adam} or AdamW~\cite{adamw} as the optimizers to obtain the optimized classification performance and have evaluated the classification performance after training each model for 20 epochs of fine-tuning procedures.

\section{Experiments}

\subsection{Experiment Setup}

\begin{figure}[htp]
    \centering
    \includegraphics[width=\columnwidth]{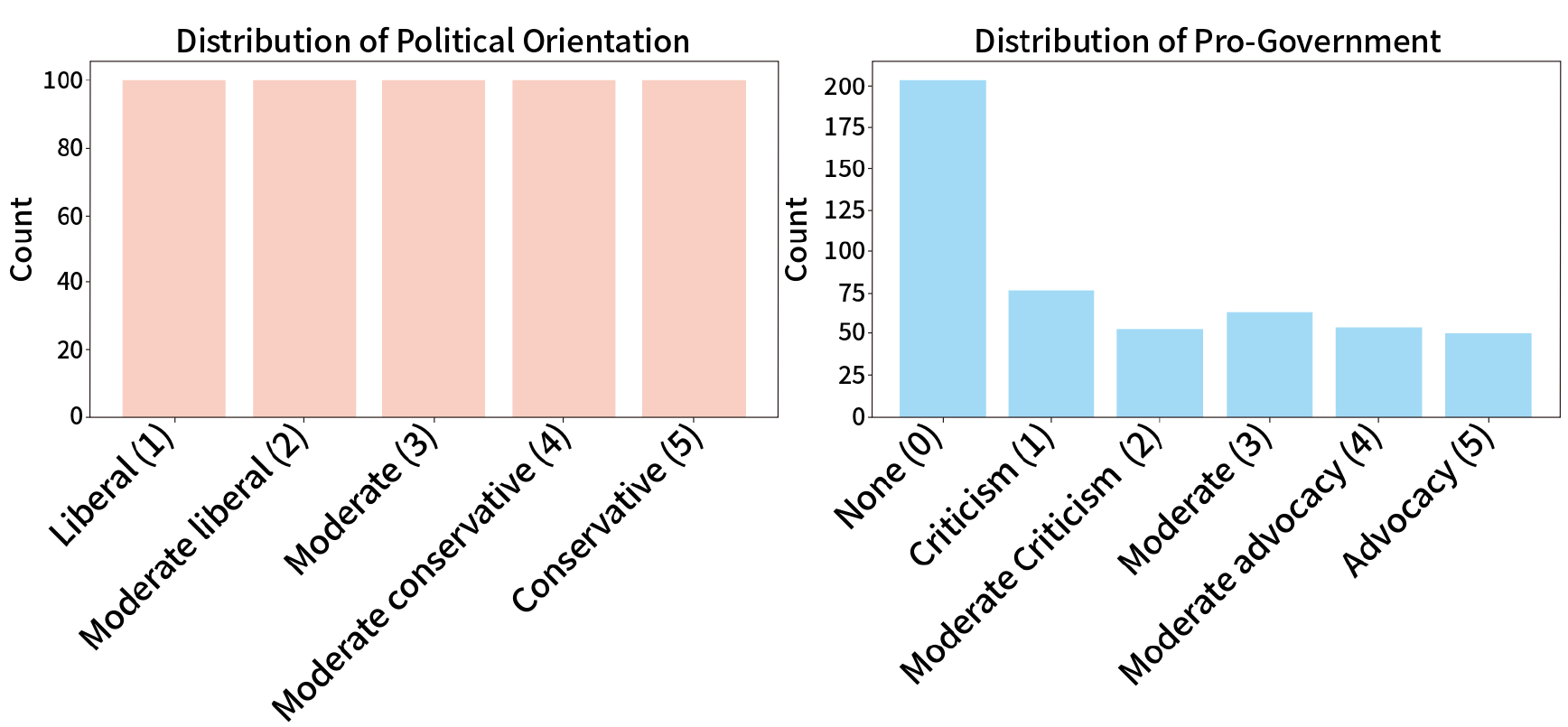}
    \caption{Class distribution of the test dataset. This figure represents the number of text samples across categories: the class distribution of (1) degrees of political orientation, and (2) the level of pro-government in our test dataset. The test dataset shows a nearly-uniform distribution.}
    \label{fig:datset_configuration3}
\end{figure}

\textbf{Data Construction.} 
We split the dataset in an 8:1:1 ratio of training, validation, and test datasets. Although the training dataset is unbalanced, we compose a test dataset that has a nearly uniform class distribution in the main experiment for reliable and consistent performance evaluation. These experimental settings can be also greatly useful in evaluating the robustness of the models for class imbalance scenarios. \\

\noindent\textbf{Baseline Architectures.} For our experiments, we use three BERT-based models pre-trained on Korean: KoBERT~\cite{kobert}, KoBigBird~\cite{kobigbird}, and KoELECTRA~\cite{koelectra}. 
We use these models from the Hugging Face which provides various Natural Language Processing (NLP) models and datasets. We conduct the tokenization using the tokenization tool in Hugging Face for each model. In particular, the KoBigBird model~\cite{kobigbird} can comprehensively process long tokens whose length is more than 2,048 because the model has a relatively large capacity and is devised to properly process long texts. Therefore, KoBigBird~\cite{kobigbird} is appropriate for analyzing articles of our dataset with tokens above 2048. We present the \textbf{KoPolitic} which adopts the KoBigBird~\cite{kobigbird} as our base architecture to leverage the multi-task approach.

\subsection{Experimental Results}

We conduct experiments with (1) the multi-task models which solve two different multi-class text classification tasks simultaneously and (2) single-task models that classify texts according to each task separately. We adopt diverse standard metrics broadly used for \textbf{KoPolitic} such as Top-2 Accuracy, Accuracy, F1-score, MAE, and Hamming Loss. Interestingly, our multi-task models, \textbf{KoPolitic}, show better classification performance compared to the single-task KoBigBird models.
Also, \textbf{KoPolitic} with 3,072 tokens, shows the best performance compared to the models adopting the shorter sequence length. 
In addition, the F1-score improves as the sequence length increases (Table~\ref{tab:result_table}), and MAE and Hamming Loss tend to decrease (Table~\ref{loss}).
This result indicates that \textbf{KoPolitic} considers the journalist's intent at the end of the article because some articles convey another opinion in the last paragraph. Therefore, \textbf{KoPolitic} with 3072 sequences tends to be effective for analyzing long articles.

\begin{figure}[h]
    \centering
    \includegraphics[width=\columnwidth]{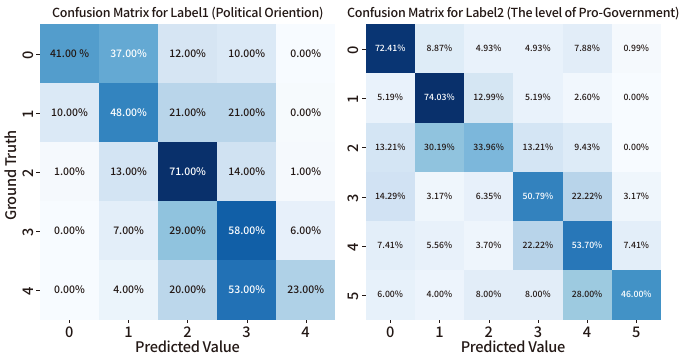}
    \caption{Confusion matrix for the task 1 (political orientation) and the task 2 (pro-government).}
    \label{confusion_matrix}
\end{figure}
\captionsetup[figure]{skip=5pt}

\noindent In the figure~\ref{confusion_matrix}, \textbf{KoPolitic} tends to perform better on moderates at 71\% on political orientation task. Moreover, we find that \textbf{KoPolitic} classify moderate liberals and moderate conservatives as moderate. This result indicates that \textbf{KoPolitic} is difficult to distinguish the subtle nuanced differences between moderate conservatives and moderate liberals. 
In the pro-government classification task, \textbf{KoPolitic} performs quite successfully on the non-target cases (class 0) at 72.41\%.
\textbf{KoPolitic} also shows better performance on strongly criticizing the government (class 1) at 74.03\% compared to advocating the government at 46\%. 
We believe criticizing words used in the article are relatively clear, whereas the advocating words are not for the classification model.

\section{Analysis}

\begin{figure*}[htp]
    \centering   \centerline{\includegraphics[width=1.2\textwidth]{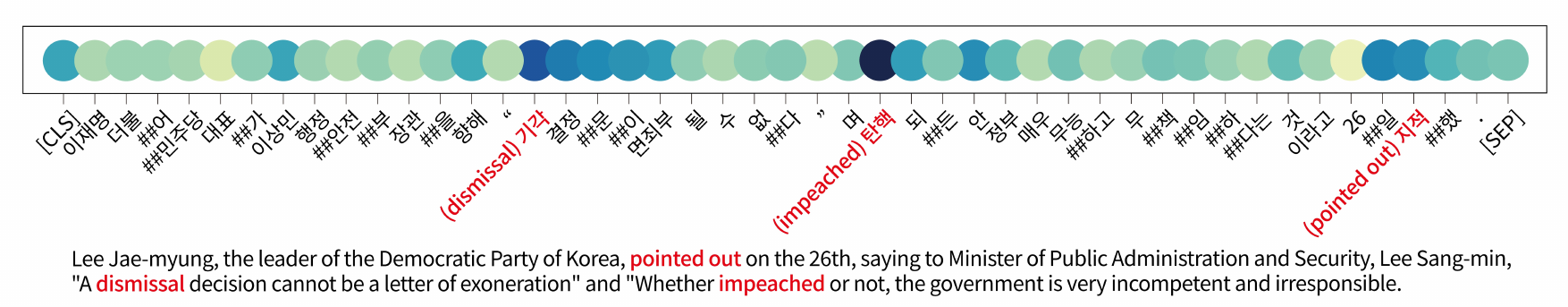}}
    \caption{Attention map for a selected sentence from the article classification results of KoPolitic. We visualize an attention map for a sentence within an article categorized under label 1 (liberal) in task 1, which represents similar nuances across various words. Notably, the terms "기각" (dismissal), "지적" (pointed out), and "탄핵" (impeachment) have garnered high attention scores.}
    \label{fig:attentionmap}
\end{figure*}

\subsection{Attention Peaks for Key Words}
Practically, calculating the attention scores for all tokens in the article might not be feasible because some texts of our dataset are too long.
Therefore, we calculate word frequency by breaking down the article into individual sentences, extracting common sentences that contain the most frequent words, and then computing the attention score. We consider that these sentences correlated with each prediction class. Moreover, we get tokens having the high attention score of common sentences on each class in Table~\ref{attention_score_table} and visualize a sample sentence in Figure~\ref{fig:attentionmap}. \\

\begin{table}[h]
    \centering
    \huge
    \renewcommand{\arraystretch}{1.6}
    \begin{adjustbox}{width=0.45\textwidth}
    \begin{tabular}{c|c|c}
        \hline
        & \multicolumn{1}{c|}{\textbf{Class}} & \multicolumn{1}{c}{\textbf{Vocabulary with High Attention Scores}} \\
        \hline
        \multirow{5}{*}{\rotatebox[origin=c]{90}{Task 1}} & 1 & Impeachment (탄핵), Rejection (기각), Blank (백지화), Incompetence (무능) \\
        \cline{2-3}
        & 2 & Defection (탈당), Responsibility (책임), Resolution (결의안), Flood (수해) \\
        \cline{2-3}
        & 3 & Summit (정상), Return (복귀), Invitation (초대), Argue (공방) \\
        \cline{2-3}
        & 4 & Use (이용), Means (수단), Weakness (약점), Instigation (선동) \\
        \cline{2-3}
        & 5 & Prevalent (만연), Moon Jae-in (문재인-Former President), War (전쟁), Fake (가짜) \\
        \hline
        \multirow{5}{*}{\rotatebox[origin=c]{90}{Task 2}} & 1 & Criticism (비판), Problem (문제), Attitude (태도), Give up (포기) \\
        \cline{2-3}
        & 2 & Point out (지적), Concern (우려), Voice (목소리), Disaster (참사) \\
        \cline{2-3}
        & 3 & Announcement (발표), Discussion (논의), Supplement (부연), Received (맞은) \\
        \cline{2-3}
        & 4 & Core (핵심), Ability (능력), Implementation (실시), Nomination (내정), Personnel (인사) \\
        \cline{2-3}
        & 5 & Definitely (반드시), Rationalization (합리화), President Yoon (윤대통령), Smash (깨부수다) \\
        \hline
    \end{tabular}
    \end{adjustbox}
    \caption{Vocabulary with high attention scores for each class. Task 1 denotes the political orientation classification, while task 2 represents the level of pro-government.}
    \label{attention_score_table}
\end{table}

\noindent\textbf{The political orientation classification (Task 1)}

\noindent We find that depending on the tokens used together, the tokens have different attention scores. 
In figure~\ref{fig:attentionmap}, strong semantic tokens like "Dismiss (기각: gigak)", "Impeach (탄핵: tanhaek)", and "Point out (지적: jijeok)" show higher attention scores.
In this sentence, the tokens such as "The name of minister (이상민: isangmin)" and "Toward (향해: hyanghae)" also show high attention scores. This result indicate that \textbf{KoPolitic} understands who the subjects of criticism are. Table~\ref{attention_score_table} shows the tokens with high attention scores for all the extracted sentences of each class. The token which describes event such as "Summit (정상 : jungsang)", "Return (복귀 : boggwi)", and "Invitation (초대 : chodae)", show high attention score.  Additionally, tokens representing mutually critical content, such as "Argue (공방 : gongbang)" also exhibit a high attention score for the class 3.

\noindent\textbf{The level of pro-government (Task 2)}

\noindent For class 1, we observe that tokens with higher attention scores are generally verbs implying criticism or negativity or nouns, the subject of criticism, rather than simply listing event tokens. "Criticism (비판: bipan)", "Problem (문제: munjae)", "Attitude (태도: taedo)", and "Give-up (포기: Pogi)" show high attention scores (Table~\ref{attention_score_table}). For class 2, verbs and specific events such as "Concern (우려: uryeo)", "Disaster (참사: chamsa) and "Voice (목소리: moksori)" tend to have a high attention score of tokens. Words related to government policy announcements, explanations, or commemorative events, for example, "Announcement (발표: balypo)", "Discussion (논의: nonyee)" and "Supplement (부연: buyeon)", exhibit high attention scores for class 3 which signifies 'Moderate'. For class 4, words related to government competence, policy implementation, and appointments, such as "Ability (능력: neunglyeog"), "Implementation (실시: silsi)", and "Personnel (인사: insa)", show high attention scores. Similarly, in the same context, words with stronger connotations, like '반드시' (necessarily), '합리화' (rationalization), and '깨부수다' (break down), show high attention scores for the class 5.

\noindent Some articles introduce the opinions mutually criticize each other with equal weight. Despite the use of the word strong tone in articles such as  "Criticism (비판: bipan)", "Objection (반대: bandae)", "Interrupt (저지: jeoji)", the result shows that \textbf{KoPolitic} can recognize the article to be "Moderate (class 3)". In conclusion, \textbf{KoPolitic} does not depend on only the highly-emphasized words. Rather than, \textbf{KoPolitic} tends to classify by considering the relationship between words.
Moreover, \textbf{KoPolitic} has the tendency to recognize the subjects clearly because \textbf{KoPolitic} is good at associating verbs with the subjects of the criticism.

\subsection{On the Real-World Distribution}

As the real world generally shows a non-uniform distribution, our dataset is also imbalanced.
However, despite the imbalance of the training dataset, we purposely set the test dataset to uniform class distribution to reduce the bias of the models. 
The evaluation of the test dataset with the real-world class distribution shows better classification performance as shown in Table ~\ref{ununiformed_results}.
Top-2 accuracy, accuracy, and F1-score consistently increase, and MAE and Hamming Loss both decrease (be improved). This suggests that when classifying article content using our model, \textbf{KoPolitic}, in the real-world setting, we can expect improved performance.

\begin{table}[h]
    \centering
    \large
    \renewcommand{\arraystretch}{1.3} 
    \begin{adjustbox}{width=0.45\textwidth}
    \begin{tabular}{c|c|c|c|c}
        \hline
        \multicolumn{2}{c|}{\textbf{Model}} & \textbf{Top-2 Accuracy} & \textbf{Accuracy} & \textbf{F1-Score} \\
        \hline
        \multirow{2}{*}{KoPolitic} & Task1 & 0.8180 & 0.6300 & 0.5530 \\
        \cline{2-5}
        & Task2 & 0.8360 & 0.6280 & 0.5406 \\
        \hline
    \end{tabular}
    \end{adjustbox}
    \caption{Classification performance of KoPolitic evaluated on the test dataset whose class distribution matches the training data distribution. The lower is better.}
    \label{ununiformed_results}
\end{table}
\vspace{-12pt}

\subsection{Single-Tasks and Multi-Task}
The single-task approaches using the KoElectra show an accuracy of 0.5120 and 0.5700, and the single-task approaches using the KoBigBird with 1024 tokens show the best Top-2 accuracy of 0.7720 and 0.8040. We believe that because a single task focuses on a specific task and can be custom-designed, a single task might be more interpretable for a specific task. We also find that these models show sub-par performance on multi-tasks, and the performance gap between single-task approaches and multi-task approaches of KoBigBird decreases as the sequence length increases. Moreover, \textbf{KoPolitic} with 3072 tokens shows better performance in multi-class than single-task model. 
This result indicates that longer sequences enable the model to understand various contexts and improve the ability of the models to generalize with transfer learning.
Furthermore, multi-task learning can prevent a model from over-fitting to a particular task and is more efficient by about 2 times compared to the single-task models in the memory and time complexity. Therefore, we believe that the classification performance of \textbf{KoPolitic} is sufficiently competitive with single-task models of KoElectra or KoBigBird.

\section{Discussion}

\textbf{Article Attributes.} In our experiments, \textbf{KoPolitic} with longer sequences generally show better classification performance. However, models of the single-task, which handle shorter tokens such as BERT or KoELECTRA, potentially show better classification performance because many articles present their primary message in the title or initial segments. However, journalists sometimes introduce an opinion of the opposite side or primary message at the end of the articles. Therefore, recognizing the whole article is necessary. Moreover, articles tend not to directly reveal the journalist's political intentions.
We find that words such as "said (말했다: malhaessda)", "did (했다: haessda)", and "be (있다: issda)" frequently appear across all labels. In Korean articles, "did (했다: haessda)" often conveys the same meaning as "said (말했다: malhaessda)". 
This suggests that journalists often include quotations from politicians in their articles. 
Using quotations can mask a journalist's true intentions, models are more difficult to predict these texts. 
This observation aligns with past studies that indicate the subtleties of political orientation are challenging to ascertain~\cite{Beyond-Binary-Labels-2017}.
Therefore, we expect future work will explore robust models, which can distinguish subtle nuanced differences.
This would be an important step toward reducing misinterpretations and improving the reliability and efficiency of natural language processing systems.

\noindent\textbf{Challenges in Korean NLP.} In the Korean language, morphology plays a pivotal role in defining grammatical significance. However, accurate and swift morphological analysis in Korean still remains a challenge. These complexities pose hurdles for Korean NLP models in understanding and interpreting every word and phrase correctly. We hope that future studies will address these challenges, enhancing the classification performance in the domain of Korean NLP fields.

\noindent\textbf{Time-Series Analysis of Political Text Data.} As our dataset only collects political articles since 2023, our dataset tends to focus on specific events and specific governments. According to the shifts in governance, subjects or contents of criticism can drastically change. 
Therefore, for future work, we expect to collect more diverse articles on different governments and to be able to compare analyses on multiple governments.

\noindent\textbf{Performance on Real-World Test Datasets.} 
In the real world, as articles that are "extremely" expressive of a particular political orientation are not common, our dataset is imbalanced. 
Therefore, future studies can utilize our dataset as a benchmark dataset to resolve the data imbalance problem because our test dataset shows a nearly-uniform class distribution in the main experiments.
For future work, we also distribute training and test datasets that are realistic distributions.

\section{Conclusion}
In this work, we introduce a new text classification dataset for recognizing the political intents in online newspapers. 
We also present a multi-task model, \textbf{KoPolitic}, as a new baseline for simultaneously classifying political orientations and the level of pro-government in the article texts. 
Since political orientation in the real world is not binary, we label our data using a 5-point scale in a multi-class classification manner.
We find some advantages of our \textbf{KoPolitic} which is a multi-task model.
First, \textbf{KoPolitic} does not just classify based on the use of strong words but also tends to understand the intent of the article by considering the relationship between the words. 
\textbf{KoPolitic} tends to focus primarily on objects and verbs when the object of criticism is clear, and in the opposite case, to focus primarily on the event. 
Second, \textbf{KoPolitic} especially tends to understand the relationship between subjects and verbs.
Therefore, \textbf{KoPolitic} tends to classify the article as moderate when journalists introduce arguments of both sides by equal weight, despite the use of several strong words. Third, \textbf{KoPolitic} with 3072 tokens shows better classification performance than the single-task of BigBird with 3072 tokens.
\textbf{KoPolitic} with 3072 tokens demonstrates the advantage of multi-task learning, such as time efficiency, resource efficiency, and the model's generalization.
Therefore, \textbf{KoPolitic} with 3072 tokens is sufficiently competitive with other models.
We hope that our research, classifying political articles that involve long tokens in a multi-task fashion, can help researchers for future work in the field of Korean Natural Language Processing (NLP).

\newpage
\nocite{*}
\section{Bibliographical References}\label{sec:reference}

\bibliographystyle{lrec-coling2024}
\bibliography{lrec-coling2024}


\end{document}